\documentclass{article}
\usepackage[preprint]{neurips_2024}
\usepackage{microtype}
\usepackage{graphicx}
\usepackage{subfigure}
\usepackage{booktabs} 
\usepackage{hyperref}

\usepackage{amsmath}
\usepackage{amssymb}
\usepackage{mathtools}
\usepackage{amsthm}
\usepackage[capitalize,noabbrev]{cleveref}
\theoremstyle{plain}

\theoremstyle{definition}

\theoremstyle{remark}

\usepackage[textsize=tiny]{todonotes}
\usepackage{booktabs}       
\usepackage{amsfonts}       
\usepackage{nicefrac}       
\usepackage{microtype}      
\usepackage{xcolor}         
\usepackage{caption}
\usepackage{subcaption}
\usepackage{amsmath,amssymb,amsfonts}
\usepackage{algorithmic}
\usepackage{graphicx}
\usepackage{textcomp}
\usepackage{xcolor}
\usepackage{tabto}
\usepackage{comment}
\usepackage{diagbox}
\usepackage{flexisym}
\usepackage{ulem}
\usepackage{multirow}
\usepackage{makecell}
\usepackage{amsmath, bm}
\usepackage{hyperref}
\usepackage{eso-pic}
\usepackage{algorithm}
\usepackage{wrapfig}
\usepackage{threeparttable}

\newcommand{\cv}[0]{\ensuremath{\boldsymbol{c}} }

\newcommand{\hv}[0]{\ensuremath{\boldsymbol{h}} }
\newcommand{\iv}[0]{\ensuremath{\boldsymbol{i}} }

\newcommand{\mv}[0]{\ensuremath{\boldsymbol{m}} }

\newcommand{\xv}[0]{\ensuremath{\boldsymbol{x}} }
\newcommand{\yv}[0]{\ensuremath{\boldsymbol{y}} }

\newcommand{\Fv}[0]{\ensuremath{\boldsymbol{F}} }

\newcommand{\Iv}[0]{\ensuremath{\boldsymbol{I}} }

\newcommand{\Mv}[0]{\ensuremath{\boldsymbol{M}} }

\newcommand{\Wv}[0]{\ensuremath{\boldsymbol{W}} }
\newcommand{\Xv}[0]{\ensuremath{\boldsymbol{X}} }

\newcommand{\Omegav}[0]{\ensuremath{\boldsymbol{\Omega}} }

\newcommand{\muv}[0]{\ensuremath{\boldsymbol{\mu}} }

\newcommand{\sigmav}[0]{\ensuremath{\boldsymbol{\sigma}} }

\usepackage[utf8]{inputenc} 
\usepackage[T1]{fontenc}    
\usepackage{hyperref}       
\usepackage{url}            
\usepackage{booktabs}       
\usepackage{amsfonts}       
\usepackage{nicefrac}       
\usepackage{microtype}      
\usepackage{xcolor}         

\title{Treating Brain-inspired Memories as Priors for Diffusion Model to Forecast Multivariate Time Series}

 \author{%
    Muyao Wang\quad
    Wenchao Chen\quad
    Zhibin Duan\quad
    Bo Chen \\
    National Laboratory of Radar Signal Processing Xidian University\\
}

\begin{document}

\maketitle

\begin{abstract}
Forecasting {\bf M}ultivariate {\bf T}ime {\bf S}eries (MTS) involves significant challenges in various application domains.
One immediate challenge is modeling temporal patterns with the finite length of the input. These temporal patterns usually involve periodic and sudden events that recur across different channels.
To better capture temporal patterns, we get inspiration from humans'  memory mechanisms and propose a channel-shared, brain-inspired memory module for MTS.
Specifically, brain-inspired memory comprises semantic and episodic memory, where the former is used to capture general patterns, such as periodic events, and the latter is employed to capture special patterns, such as sudden events, respectively.
Meanwhile, we design corresponding recall and update mechanisms to better utilize these patterns.
{Furthermore, acknowledging the capacity of diffusion models to leverage memory as a prior, we present a brain-inspired memory-augmented diffusion model. This innovative model retrieves relevant memories for different channels, utilizing them as distinct priors for MTS predictions. This incorporation significantly enhances the accuracy and robustness of predictions.}
Experimental results on eight datasets consistently validate the superiority of our approach in capturing and leveraging diverse recurrent temporal patterns across different channels.
\end{abstract}

\section{Introduction}
Multivariate time series (MTS) plays an important role in a wide variety of domains, including internet services~\citep{dai2021sdfvae}
industrial devices~\citep{finn2016unsupervised,oh2015action}
, health care~\citep{choi2016retain,choi2016doctor}, finance~\citep{maeda2019effectiveness,gu2020price}
and so on. 
With the development of neural networks, deep
learning models, such as  CNN-based models \citep{wu2022timesnet}, Transformer-based models \citep{liu2023itransformer, zhou2022fedformer, zhou2021informer} and Linear-based models \citep{zeng2023transformers} have been applied to MTS and achieved better performance compared with
traditional statistical methods \citep{box2015time,ariyo2014stock} in MTS forecasting tasks in terms of their higher capacity. Besides the advanced design of model architectures, some researchers have focused on diffusion theory for MTS forecasting and achieved promising results recently \citep{shen2024multiresolution, pmlr-v202-shen23d, rasul2021autoregressive, alcaraz2022diffusion, tashiro2021csdi}.

\begin{wrapfigure}{!r}{7cm}\vspace{-1mm}
    \centerline{\includegraphics[width=1\linewidth]{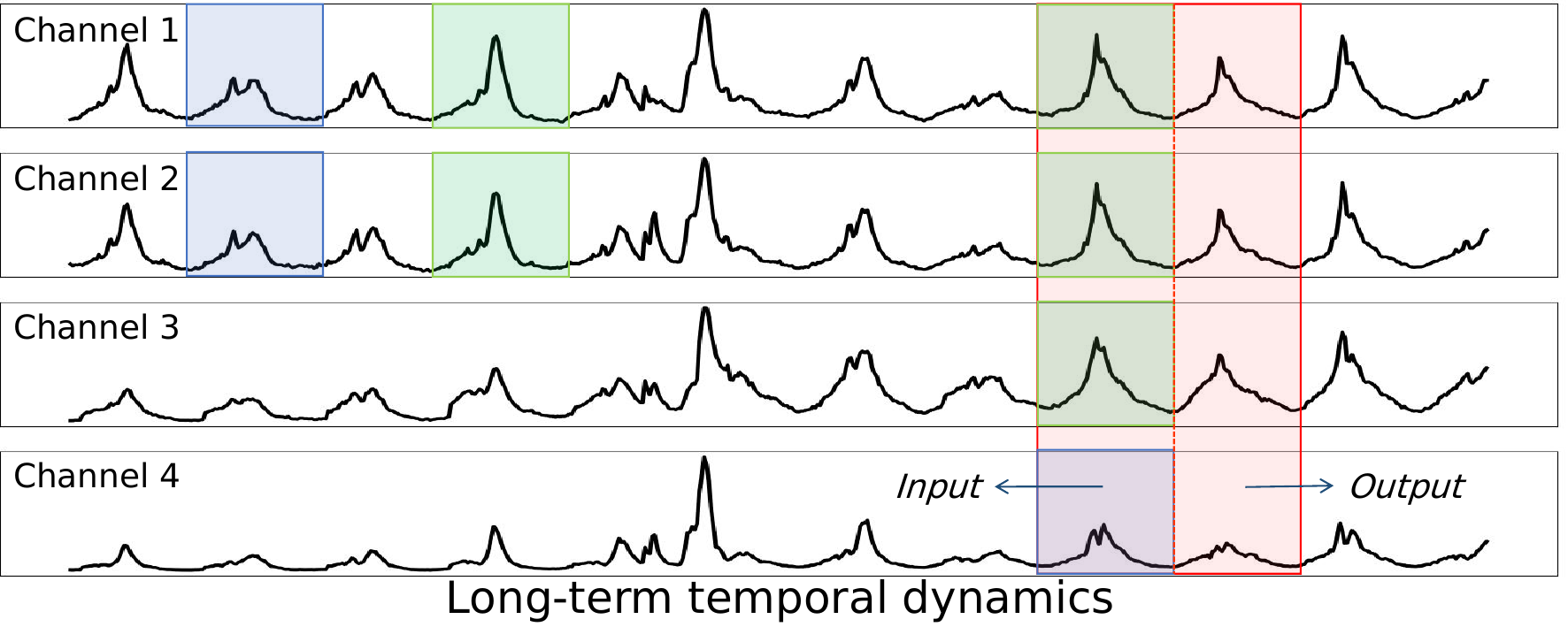}}
	\caption{The red box represents the current MTS forecasting task. The blue box and green box represent diverse temporal patterns across different channels which can help channel 3 and 4 to forecast better. Because those temporal patterns have happened in other channels in history.} 
    \label{motivation}
\end{wrapfigure}

Despite the remarkable progress achieved by employing advanced methods \citep{nie2022time,zeng2023transformers,shen2024multiresolution} to explore long-range temporal dependencies in MTS, predicting future observations over long time steps from limited past examples remains a challenging problem. 
{{One contributing factor to this challenge is the temporal information is extracted solely from the lookback window and not from the entire time series.}}
Similar challenges are encountered in language modeling, where model performance is constrained by context length \citep{yang2020retransformer,wang2020linformer}. Substantial efforts have been invested in developing external knowledge modules \citep{bulatov2022recurrent, guu2020retrieval, gao2023retrieval}, such as memory module, for language modeling to extend context information.

While memory modules have shown effective progress in Natural Language Processing (NLP), to the best of our knowledge, few studies have explored their adaptation to MTS forecasting tasks.
Adapting a memory module to MTS is not straightforward, mainly because multivariate time series typically encompass multiple channels with specific connections among them \citep{nie2022time, liu2023itransformer}.
As exemplified in Fig. \ref{motivation} through the green and blue boxes, it is evident that the same temporal patterns can recur in different channels.
To effectively capture the potential recurrent characteristics of temporal patterns across different channels, the designed MTS memory module should be shared among different channels. Simultaneously, temporal patterns in MTS can often be categorized into general patterns and special patterns, as illustrated in Fig.\ref{motivation}. The former, such as cyclic variations, tends to occur more frequently, while the latter, such as sudden events, occurs less frequently.

To simultaneously capture both general patterns and special patterns in time series, we draw inspiration from brain memory \citep{tulving1972episodic,renoult2012personal} to formulate a brain-inspired memory module. In line with the structure of brain memory, we construct two components within our memory module. The first component is semantic memory storing the knowledge information in the human brain, which learns the general temporal patterns between different channels, enabling the network to exhibit human-like generalization abilities across different channels.
The second component is episodic memory storing the representative events in the human brain, which stores the special temporal patterns across different channels that are difficult to predict, allowing the network to remember important exceptions. Subsequently, we introduce two pivotal processes in the brain-inspired memory module: 1) memory recall, which retrieves the relevant temporal patterns from brain-inspired memory to apply to the MTS forecasting task; 2) memory update, which continuously summarizes general patterns and selects special patterns during the training process and gradually consolidates the memories stored in brain-inspired memory.

In addition, the Bayesian generative model can not only naturally incorporate these memories as prior knowledge into the generation process, but also model the uncertainty in temporal patterns caused by noise \citep{han2022card,salinas2020deepar,wang2024considering}.
Given the promising results achieved by diffusion models in MTS forecasting \citep{shen2024multiresolution}, we advocate using a diffusion model to encompass this non-deterministic temporal information, thereby enhancing predictive performance across various channels.
Once our network possesses brain-inspired memory akin to human cognition, we can leverage them as priors for a diffusion model.
By doing so, we can construct a generative model named {\bf B}rain-{\bf i}nspired {\bf m}emory-augmented {\bf Diff}usion Model ({\bf Bim-Diff}), effectively modeling general and special temporal patterns.
The main contributions of our work are summarized as follows:
\begin{itemize} 
    \item We propose a brain-inspired memory consisting of semantic and episodic memory, which can capture general and special temporal patterns across different channels within the MTS.
    \item We design efficient update and recall mechanisms for the proposed memory module, enabling it to capture and utilize two distinct types of temporal patterns effectively.
    \item We develop Bim-Diff, an efficient probabilistic model that integrates the two memories into the diffusion framework as the conditional prior, for MTS forecasting.
    \item Experiments on eight real-world datasets illustrate the efficiency of our model on the MTS forecasting task. Specifically, Bim-Diff ranks  top-1 under the average ranking setting among the advanced models for comparison in the main experiment.
\end{itemize}

\section{Related Work}
In recent decades, the field of MTS forecasting has evolved significantly. It has transitioned from conventional statistical approaches such as ARIMA \citep{ariyo2014stock} and machine learning techniques like GBRT \citep{friedman2001greedy} towards more advanced deep learning-based solutions, including Recurrent Neural Networks \citep{lai2018modeling}, Convolutional Neural Networks \citep{bai2018empirical}, \citep{liu2021time} and Multi-layer Perceptron based models \citep{zeng2023transformers, oreshkin2019n,challu2023nhits}. 
More recently, several Transformer-based models have demonstrated remarkable efficacy in capturing temporal dependencies for MTS forecasting. Notable examples include Autoformer \citep{wu2021autoformer}, which proposes a decomposition architecture with an Auto-Correlation mechanism to capture long-range temporal dependencies, and FEDformer \citep{zhou2022fedformer}, which utilizes a seasonal-trend decomposition with frequency-enhanced blocks for the same purpose. PatchTST \citep{nie2022time} performs time series prediction by patching the time series and making different channels to extract local semantic information independently. iTransformer \citep{liu2023itransformer} has recently achieved promising performance by simply inverting the vanilla Transformer to capture the channel relationships.

Very recently, diffusion models have also been developed for time series data. TimeGrad \citep{rasul2021autoregressive} uses a recurrent neural network as a condition to predict in an autoregressive manner. Due to its prediction manner suffers from slow inference on long time series, CSDI \citep{tashiro2021csdi} proposes a non-regressive method to solve the problem. Besides, SSSD \citep{alcaraz2022diffusion} adjusts the mode structure and further accelerates the inference speed of the diffusion model. As SSSD and CSDI are mask-based condition methods, which suffer from the problem of boundary disharmony. To solve the problem, the non-autoregressive TimeDiff method \citep{pmlr-v202-shen23d} uses future mixup and autoregressive initialization for conditioning. To make the diffusion model leverage the time series characteristic better, mrDiff \citep{shen2024multiresolution} proposes a multi-resolution temporal structure and achieves state-of-the-art.
However, due to the uncertainty in MTS forecasting, as well as challenges related to limited input length and long prediction horizons, overall, MTS forecasting remains a significant challenge.

\section{Memory Augmented Diffusion Model}

\subsection{Problem Definition}

Defining the MTS as $\xv=\{\xv_{1}, \xv_{2}, ..., \xv_{L}\}$ , where $L$ is the duration of $\xv$ and the observation at time $t$, $\xv_{t} \in \mathbb{R}^{N}$, where $N$ denotes the number of channels, thus $\xv \in \mathbb{R}^{L \times N}$.  multivariate time series forecasting aims to predict the value of $\yv_{1:H}$, where $H$ is the number of time steps in the future.

\subsection{Denoising Diffusion Probabilistic Models}
The Denoising Diffusion Probabilistic Model (DDPM)\citep{ho2020denoising} is a kind of well-known probabilistic model, which usually consists of a fixed forward process that can map a complex data distribution to an easy prior distribution, such as a normal standard distribution, and a learnable backward denoising process that can generate a sample from the complex data distribution. We specify the forward process conditional distributions as:
\begin{equation} \textstyle \label{forwardprocess}
\begin{array}{l}
q\left(\xv^{k} \mid \xv^{k-1}\right)=\mathcal{N}\left(\xv^{k} ; \sqrt{1-\beta_{k}} \xv^{k-1}, \beta_{k} \Iv\right), \quad k=1, \ldots, K
\end{array}
\end{equation}
where at the $k$th step, $x^k$ is generated by corrupting the previous iterate $x^{k-1}$(scaled by $\sqrt{1-\beta_k}$) with standard normal distribution(with variance $\beta_k \in [0,1]$).
The formula \eqref{forwardprocess} admits a closed-form sampling distribution with an arbitrary step k: $q\left(\xv^{k} \mid \xv^{0}\right)=\mathcal{N}\left(\xv^{k}; \sqrt{\overline{\alpha}_k} \xv^{0}, (1-\bar{\alpha}_{k}) \Iv\right)$, where
$\bar{\alpha}_{k}=\prod_{s=1}^k\alpha_s$ and $\alpha_k=1-\beta_k$.
Such formula corresponds to a tractable forward process posterior:
\begin{equation} \textstyle \label{forwardposterior}
\begin{array}{l}
q\left(\xv^{k-1} \mid \xv^{k}, \xv^0\right)=\mathcal{N}\left(\xv^{k-1} ; \Tilde{\mu}(\xv^k,\xv^0), \Tilde{\beta}_{k} \Iv \right)
\end{array}
\end{equation}
where $\Tilde{\mu}:=\frac{\sqrt{\alpha_{k}}\left(1-\bar{\alpha}_{k-1}\right)}{1-\bar{\alpha}_{k}} \xv^{k}+\frac{\sqrt{\bar{\alpha}_{k-1}} \beta_{k}}{1-\bar{\alpha}_{k}} \xv^0, \Tilde{\beta}_{k}:=\frac{\left(1-\bar{\alpha}_{k-1}\right)}{\left(1-\bar{\alpha}_{k}\right)}\beta_k$.

In DDPM\citep{ho2020denoising}, backward denoising is defined as a Markovian process. Specifically, at the $k$th denoising step, $x^{k-1}$ can be sampled from the following conditional distribution:
\begin{equation} \textstyle \label{backwarddenoise}
\begin{array}{l}
p_\theta\left(\xv^{k-1} \mid \xv^{k} \right)=\mathcal{N}\left(\xv^{k-1} ; {\mu_\theta}(\xv^k,k), \sum_\theta(\xv^k,k) \right), \quad k=1, \ldots, K
\end{array}
\end{equation}
where the variance $\sum_\theta(\xv^k,k)$ is usually fixed the same as $\Tilde{\beta}_{k}$ in Eq.\eqref{forwardposterior}. Thus, to approximate the tractable forward process posterior, parameter $\theta$ is learned by minimizing the loss $\mathcal{L}=\left\|\mu_\theta(\xv^k,k)-\Tilde{\mu}\right\|^2$. According to the \citep{ho2020denoising}, the final loss function can be defined as:
\begin{equation} \textstyle \label{lossfunction}
\begin{array}{l}
\mathcal{L}=\mbox{E}_{\xv^0,\epsilon,k}\left\|\xv^0-\xv_\theta(\xv^k, k)\right\|^2
\end{array}
\end{equation}
which means uses a network $\xv_\theta$ to obtain an estimate $\xv_\theta(\xv^k,k)$ of the clean data $\xv^0$ given $\xv^k$, the $\epsilon$ represents a random noise sampled from the normal standard distribution.

\subsection{Memory-augmented Conditional Diffusion Models for Time Series Prediction}
{\bf Theory framework:} When using conditional diffusion models for time series prediction, the following distribution is considered \citep{shen2024multiresolution,rasul2021autoregressive}.
\begin{equation} \textstyle \label{conditionalbackward}
\begin{array}{l}
p_{\theta}\left(\yv_{1: H}^{0: K} \mid \cv \right)=p_{\theta}\left(\yv_{1: H}^{K}\right) \prod_{k=1}^{K} p_{\theta}\left(\yv_{1: H}^{k-1} \mid \yv_{1: H}^{k}, \cv \right), \cv=\mathcal{F}\left(\xv_{1:L}^{0}\right)
\end{array}
\end{equation}
where $\yv^K_{1:H}\sim\mathcal{N}(0,1)$, c is the condition, and $\mathcal{F}$ is a conditioning network that takes the past observations $\xv_{1:L}^0$ as input. Different from the deterministic prior $\cv$ used in the previous method \citep{shen2024multiresolution}, we propose a variational brain-inspired memory $\mv$ as diffusion model priors to enhance the model performance. Thus, the prediction process of $\yv^0_{1:H}$ can be defined as follow:

\begin{equation} \textstyle \label{predictionprocess}
\begin{array}{l}
p_{\theta}\left(\yv_{1: H}^{0}\right)=\iiint p_{\theta}\left(\yv_{1: H}^{0:K} \mid \cv\right)p_\theta(\cv \mid \mv)p(\mv \mid \xv^0_{1:L}, \Mv)  d\cv d\mv d\yv_{1:H}^{1:K}
\end{array}
\end{equation}

The condition vector $\cv$ can be sampled from a condition distribution $p_\theta(\cv \mid \mv)=\mathcal{N}(\muv(\mv), \mbox{diag}(\sigmav(\cdot))$ and the $\mv$ can be sampled from the conditional distribution: $q_{\phi}(\mv \mid \xv^0_{1:L}, \Mv)=\mathcal{N}(\muv(\mv^e,\mv^s), \mbox{diag}(\sigmav(\cdot))$. Specifically, $\muv(\mv) = \Wv_1(\mv), \muv(\mv^e,\mv^s) = \Wv_2(\mv^e+\mv^s)$, where $\Wv_1, \Wv_2 \in R^{L\times L}$ are learnable matrixes, and $\mv^e, \mv^s$ are recalled from memory blocks $\Mv$ by Eq.\eqref{memorye}, Eq.\eqref{memoryk} respectively, which will be introduced in the next section in detail, and $\mbox{diag}(\sigmav(\cdot))$ is a learnable diagonal matrix $\Iv$.

{\bf Conditional denoising network:} Inspired by \citep{pmlr-v202-shen23d}, we use the same future mixup strategy to produce the conditioning signal to guide the training as follows:
\begin{equation} \textstyle \label{mixup}
\begin{array}{l}
\cv_{\text {mix }}=\mv^{mask} \odot \cv+\left(1-\mv^{mask}\right) \odot \yv_{1: H}^{0}
\end{array}
\end{equation}
where $\mv^{mask} \in [0, 1)^{H\times N}$  is a mixing matrix with each element randomly sampled from the uniform distribution on $[0, 1)$, and $\odot$ is the Hadamard product.

Analogous to Eq.\eqref{backwarddenoise}, the conditional denoising process at step k is denied by:
\begin{equation} \textstyle \label{conditionbackwarddenoise}
\begin{array}{l}
p_\theta\left(\yv^{k-1}_{1:H} \mid \yv^{k}_{1:H} \right)=\mathcal{N}\left(\yv^{k-1}_{1:H} ; {\mu_\theta}(\yv^k_{1:H},k,\cv_{\text{mix}}), \Tilde{\beta}_k\Iv \right), \quad k=1, \ldots, K
\end{array}
\end{equation}
where ${\muv_\theta}:=\frac{\sqrt{\alpha_{k}}\left(1-\bar{\alpha}_{k-1}\right)}{1-\bar{\alpha}_{k}} \yv^{k}_{1:H}+\frac{\sqrt{\bar{\alpha}_{k-1}} \beta_{k}}{1-\bar{\alpha}_{k}} \yv_{\theta}(\yv^k_{1:H},k,c_{\text{mix}})$, $\yv_{\theta}(\yv^k_{1:H},k,c_{\text{mix}})$ is an estimation of input $\yv_{1:H}^0$, and $\theta$ includes all parameters in the conditional denoising network. Finally, analogous to Eq.\eqref{lossfunction}, $\theta$ is obtained by minimizing the following denoising objective:
\begin{equation} \textstyle \label{conditionlossfunction}
\begin{array}{l}
\mathcal{L_{\text{condition}}}=\mbox{E}_{\yv^0_{1:H},\epsilon,k}\left\|\yv^0_{1:H}-\yv_{\theta}(\yv^k_{1:H},k,\cv_{\text{mix}})\right\|^2
\end{array}
\end{equation}
On inference, the ground truth $\yv^0_{1:H}$ is no longer available. Hence, we simply set $c_{\text{mix}}=c$ in Eq.\eqref{mixup}. Based on the denoise process Eq.\eqref{conditionbackwarddenoise}, the final prediction $\hat{\yv}_{1:H}$ can be obtained from random noise following the below formula step by step:
\begin{equation} \textstyle \label{getpredict}
\begin{array}{l}
\hat{\yv}^{k-1}_{1:H}=\frac{\sqrt{\alpha_{k}}\left(1-\bar{\alpha}_{k-1}\right)}{1-\bar{\alpha}_{k}} \hat{\yv}^{k}_{1:H}+\frac{\sqrt{\bar{\alpha}_{k-1}} \beta_{k}}{1-\bar{\alpha}_{k}} \yv_{\theta}(\hat{\yv}^k_{1:H},k,\cv_{\text{mix}})+\Tilde{\beta}_k\epsilon, \quad k=1, \ldots, K
\end{array}
\end{equation}
where the $\epsilon$ represents a random noise sampled from the normal standard distribution.

\section{Capturing Temporal Patterns via Brain-inspired Memory}
\begin{figure*}
\includegraphics[width=1\linewidth]{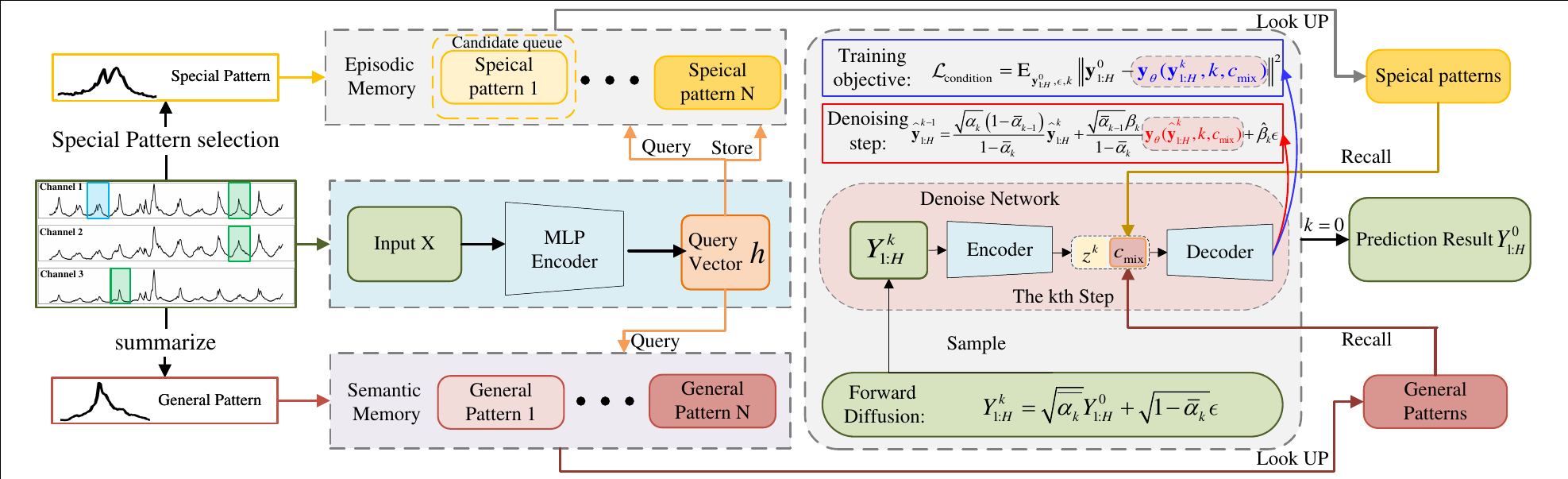}
\caption{{The framework of brain-inspired memory-augmented diffusion model.}}\label{fig:framework} 
\end{figure*}

In this section, we will introduce Brain-inspired memory (Bim), equipped with the mechanism of memory recall and update strategy. The fundamental concept behind Bim draws inspiration from the memory mechanisms observed in the human brain \citep{tulving1972episodic, renoult2012personal}: 1) there are two advanced forms of memory in the human brain, specifically, semantic memory allows the summary of general conceptual information, and episodic memory allows the collection of detailed events; 2) In the task of forecasting MTS, the semantic memory can summary the general temporal patterns of the MTS, while the episodic memory can store the special temporal patterns of the MTS as we mentioned before.

\subsection{{{Semantic Memory}}}

As shown in the Fig~\ref{fig:framework}, there are $N_{1}$ learnable semantic memory blocks in semantic memory module denoted as ${\Mv^{s} = \left\{\Mv^{s}_{i} \right\}_{i=1}^{N_1} }$, where $i$ denotes the $i$th block and $s$ means the semantic memory.
The purpose of designing semantic memory is to encapsulate and summarize the recurrent general patterns across different channels in the historical series.

{\bf Storage strategy:} The initialization of the semantic memory is random, and the memory is learnable during the training process. To summarize general patterns across different channels, each semantic memory block is shared among different channels and is specifically designed to store univariate temporal patterns referred to as general patterns. In line with our earlier discussion, we initialize the $i$-th block (pattern) of semantic memory with a learnable vector $\Mv^{s}_{i} \in T \times 1$, sampled from a normal Gaussian distribution.

{\bf Recall strategy:} For the MTS forecasting task, the semantic memory initially computes the attention score between each channel and each general pattern stored in the semantic memory. Subsequently, an attention mechanism is employed to aggregate these patterns into a semantic memory $\mv^{s}_{i}$ for predicting future values. The attention score can be calculated as follows:
\begin{equation} \textstyle \label{attentionscorek}
\small
\begin{array}{l}
\mbox{score}\left( \Mv_i^s, \hv_j\right)=\mbox{Cosine}\left( \Mv_i^s, \hv_j\right), \mbox{Cosine}\left( \Mv_i^s, \hv_j\right) = \frac{\Mv_i^s\hv_j}{\|\Mv_i^s\|\|\hv_j\|}
\end{array}
\end{equation}
where $\mbox{score}(\cdot)$ means the attention score function defined by cosine similarity; $\hv_j$ denotes the $j$th channel of query vector $\hv$ which is calculated by an MLP temporal encoder as shown in Fig~\ref{fig:framework}; 
Then, to get the semantic memory for the MTS, we can further aggregate these general patterns with their attention weights, formulated as:
\begin{equation} \textstyle \label{memoryk}
\begin{array}{l}
{\mv}_{j}^{s}=\sum_{i=1}^{N_1} \frac{\mbox{score}\left({\Mv}_{i}^{s}, {\hv}_{j}\right)}{\sum_{i=1}^{N_1} \mbox{score}\left({\Mv}_{i}^{s}, {\hv}_{j}\right)} {\Mv}_{i}^{s}
\end{array}
\end{equation}
where $\mv_j^s \in \mbox{R}^{T\times1}$ has gathered the general patterns information from the semantic memory for the jth channel of MTS.

{\bf Update strategy:} As we mentioned before, the semantic memory is calculated by general patterns stored in the semantic memory module. When we predict the future value, each learnable general pattern within the semantic memory takes into account the prediction results of each channel in the MTS during gradient backpropagation.  Such an approach effectively accomplishes the objective of identifying general patterns across different channels. Furthermore, to keep memory items as compact as possible, at the same time as dissimilar as possible, we use two constraints \citep{gong2019memorizing} \citep{jiang2023spatio}, including a consistency loss L1 and a contrastive loss L2, denoted by
\begin{equation} \textstyle \label{lamda}
\small
\begin{array}{l}
{L}_{1}=\sum_{j}^{N}\left\|\hv_{j}-\Mv_j^{s, 1}\right\|^{2} \\
{L}_{2}=\sum_{j}^{N} \max \left\{\left\|\hv_{j}-\Mv_j^{s, 1}\right\|^{2}-\left\|\hv_{j}-\Mv_j^{s, 2}\right\|^{2}+\lambda, 0\right\}
\end{array}
\end{equation}
where $\Mv_j^{k, i} \in \mbox{R}^{T\times1}$ denotes the ith similar semantic pattern of channel j; $\lambda$ denotes the margin between the first similar general pattern and the second similar general pattern; N indicates the number of MTS channels.
More details about the update strategy are described in Appendix \ref{B.1.2}

\subsection{{{Episodic Memory}}}
\begin{algorithm} [t]
\small
\caption{\small The update strategy of episodic memory}
\label{algorithm1}
\begin{algorithmic}
\STATE Set mini-batch size as $N_{batch}$;
 \STATE Initialize the candidate queue as $\left\{Q^{e}_{i} \right\}_{N_3}$, the episodic memory as $\left\{M^{e}_{i} \right\}_{N_2}$, the frequency matrix$F_{M_i^e}$;
 \STATE Use {\bf{select}} function Eq.\eqref{initiale} to select $k$ special patterns from the mini-batch until the episodic memory is full;
 \REPEAT 
 \STATE Sort the patterns by frequency matrix $F_{M_i^e}$ in the episodic memory $\left\{M^{e}_{i} \right\}_{N_2}$ and the last $k$ element in the candidate queue $\left\{Q^{e}_{i} \right\}_{N_3}$ together;
 \STATE Pop out the last $k$ elements in the candidate queue;
 \STATE Pop in the new $k$ special patterns selected from the current mini-batch.
\UNTIL{no more new patterns}
\STATE return global episodic memory $\left\{M^{e}_{i} \right\}_{N_2}$.
\end{algorithmic}
\end{algorithm}

As shown in the Fig~\ref{fig:framework}, there are $N_{2}$ episodic memory blocks in episodic memory module denoted as ${M^{e} = \left\{M^{e}_{i} \right\}_{N_2} }$, where $i$ denotes the $i$th block and $e$ means the episodic memory. Unlike semantic memory, episodic memory tends to focus on storing more special patterns within MTS directly, meaning no learnable parameters are included in it.

{\bf Storage strategy:}
The design of the episodic memory draws inspiration from prior works \citep{fortunato2019generalization, guo2020improved}. The key concept is to utilize a memory module to gather a subset of representative data samples. Specifically, the $i$-th block (pattern) of the episodic memory is a vector of univariate series selected from the query vector of MTS, denoted as $\hv$, and is named special pattern, which has been encountered during training:
\begin{equation} \textstyle \label{initiale}
\small
\begin{array}{l}
\Mv^{e}_{i} = \mbox{select}(\hv) \\
\end{array}
\end{equation}
where $\mbox{select}(\cdot)$ means the select function which is defined in update strategy; $\hv$ denotes the latent feature of past MTS. The episodic memory is empty at first because it has never seen any MTS.

{\bf Recall strategy:} Similar to the recall method in semantic memory, we first calculate the attention score according to the query vector $\hv_j$. Then we select top-k similar results for aggregating steps, which is a little different from the semantic memory. It is because semantic memory aims to learn the general patterns during the training process, while episodic memory is designed to focus on finding the special patterns that can be referenced across different channels.
The detail of recalling episodic memory is defined as follows:
\begin{equation} \textstyle \label{memorye}
\begin{array}{l}
{\mv}_{j}^{e}= \sum_{i=1}^{K} \frac{\mbox{score}\left({\Mv}_{i}^{e}, {\hv}_{j}\right)}{\sum_{i=1}^{K} \mbox{score}\left({\Mv}_{i}^{e}, {\hv}_{j}\right)} {\Mv}_{i}^{e}
\end{array}
\end{equation}
the episodic memory $\mv_j^e \in \mbox{R}^{T \times 1}$ has gathered the historical special event information through recalling the episodic memory $\Mv^e$ for the jth channel.

{\bf Update strategy:}
Inspired by the fact that it is important for the human brain to remember those special examples he/she often recalls, we design a special pattern selection and frequency-based episodic memory update strategy. 

{\textbf{$\iv)$} \bf{special pattern selection:}} When a batch of MTS is presented during training, we compute the loss of each MTS in the batch. Subsequently, we select the MTS with the largest loss as special examples. Each special example is then divided into $n$ univariate series, denoted as special patterns, which are stored in $n$ episodic memory blocks ${\{\Mv_i^e\}}_n$, where $n$ represents the number of channels in each MTS. This process defines the select function Eq.\eqref{initiale} mentioned earlier.

{\bf{$\iv\iv)$ frequency-based episodic memory update strategy}}: It is acknowledged that the most frequently accessed special pattern is the most representative.
In order to record how many times $\Mv^e_i$ was recalled during training, we use a frequency matrix denoted as $\Fv_{M_{i}^e}$. 
Traditionally, $\Mv_i^e$ is updated by replacing the least used pattern in the block if its capacity exceeds the limit $N_2$. However, since newly incoming patterns have a lower access frequency rate than previously incoming patterns, these new patterns are more likely to be replaced, and we define this issue as memory freshness degradation.

To address this issue, we introduce a circular candidate queue within the episodic memory, designed to store the special patterns from the past $N_3$ occurrences, denoted as ${Q^{e} = \left\{Q^{e}_{i} \right\}_{N_3} }$, with the constraint $N_3 \leq N_2$. 
It's worth noting that the frequency matrix $\Fv_{M_{i}^e}$ is reset to a zero matrix following each update of the episodic memory. The detailed memory updating process algorithm is described in Algorithm \ref{algorithm1}. More details can be found in Appendix \ref{B.1.1}.
\section{Model Training}
Similar to conditional diffusion models, the optimization objective of our model can be computed as 
\begin{equation} \label{ELBO} \textstyle
\begin{split}
\begin{array}{l}
Loss = \mathcal{L_{\text{condition}}}+\alpha_1 L_1+\alpha_2 L_2
\end{array}
\end{split}
\end{equation}
where $\mathcal{L_{\text{condition}}}$ is defined by Eq.\eqref{conditionlossfunction}, $L_1, L_2$ is computed by Eq.\eqref{lamda}; {$\alpha_1, \alpha_2$} indicates the balance parameter of two constraints, which is chosen by grid search on the validation set.

\begin{table*}[!t]
\caption{Multivariate prediction MAEs on the real-world time series datasets (subscript is the rank). CSDI runs out of memory on Electricity. Results of all baselines are from \citep{shen2024multiresolution}.}\vspace{-2mm}
	\label{tab:forecasting_result}
	\begin{center}
        \setlength{\tabcolsep}{1.0mm}
         \resizebox{1\hsize}{!}{
\begin{tabular}{c|cccccccc|c}
\toprule
            & NorPool         & Caiso         & Electricity        & Weather         & Exchange         & ETTh1         & ETTm1         & Wind          & avg rank \\ \midrule
Bim-Diff    & \uline{0.604}$\pm0.001_{(2)}$ & {\bf0.193}$\pm0.002_{(1)}$ & 0.248$\pm0.001_{(3)}$      & \uline{0.320}$\pm0.002_{(2)}$   & {\bf 0.079}$\pm0.001_{(1)}$ & {\bf0.413}$\pm{0.002}_{(1)}$ & {\bf0.361}$\pm0.001_{(1)}$ & {\bf0.673}$\pm0.002_{(1)}$ &   \bf{1.5}       \\ \midrule
mr-Diff     & \uline{0.604}$_{(2)}$   & 0.219$_{(5)}$ & 0.252$_{(4)}$      & 0.324$_{(3)}$    & 0.082$_{(4)}$    & 0.422$_{(3)}$ & 0.373$_{(3)}$ & \uline{0.675}$_{(2)}$ &    \uline{3.3}      \\
TimeDiff    & 0.611$_{(4)}$   & 0.234$_{(8)}$ & 0.305$_{(7)}$      & \bf{0.312}$_{(1)}$    & 0.091$_{(9)}$    & 0.430$_{(4)}$ & \uline{0.372}$_{(2)}$ & 0.687$_{(4)}$ &     4.9     \\
TimeGrad    & 0.821$_{(20)}$   & 0.339$_{(19)}$ & 0.630$_{(22)}$      & 0.381$_{(16)}$   & 0.193$_{(21)}$    & 0.719$_{(23)}$ & 0.605$_{(23)}$ & 0.793$_{(23)}$ &    20.9      \\
CSDI        & 0.777$_{(18)}$   & 0.345$_{(20)}$ & -                  & 0.374$_{(14)}$   & 0.194$_{(22)}$    & 0.438$_{(7)}$ & 0.442$_{(17)}$ & 0.741$_{(12)}$ &     15.7     \\
SSSD        & 0.753$_{(15)}$   & 0.295$_{(12)}$ & 0.363$_{(13)}$      & 0.350$_{(10)}$   & 0.127$_{(18)}$    & 0.561$_{(19)}$ & 0.406$_{(12)}$ & 0.778$_{(20)}$ &     13.6     \\ \midrule
$D^3$VAE    & 0.692$_{(11)}$   & 0.331$_{(17)}$ & 0.372$_{(15)}$      & 0.380$_{(15)}$   & 0.301$_{(23)}$    & 0.502$_{(16)}$ & 0.391$_{(10)}$ & 0.779$_{(21)}$ &    16.0      \\
CPF         & 0.889$_{(22)}$   & 0.424$_{(22)}$ & 0.643$_{(23)}$      & 0.781$_{(24)}$   & 0.082$_{(4)}$    & 0.597$_{(21)}$ & 0.472$_{(18)}$ & 0.757$_{(17)}$ &    18.8      \\
PSA-GAN     & 0.890$_{(23)}$   & 0.477$_{(23)}$ & 0.533$_{(21)}$      & 0.578$_{(23)}$   & 0.087$_{(8)}$    & 0.546$_{(18)}$ & 0.488$_{(19)}$ & 0.756$_{(15)}$ &     18.8     \\ \midrule
N-Hits      & 0.643$_{(8)}$   & 0.221$_{(6)}$ & \uline{0.245}$_{(2)}$      & 0.335$_{(5)}$   & 0.085$_{(6)}$    & 0.480$_{(10)}$ & 0.388$_{(8)}$ & 0.734$_{(10)}$ &      6.9    \\
FiLM        & 0.646$_{(10)}$   & 0.278$_{(10)}$ & 0.320$_{(9)}$      & 0.336$_{(6)}$   & {\bf0.079}$_{(1)}$    & 0.436$_{(6)}$ & 0.374$_{(4)}$ & 0.717$_{(7)}$ &   6.6       \\
Depts       & 0.611$_{(4)}$   & 0.204$_{(3)}$ & 0.401$_{(20)}$      & 0.394$_{(18)}$   & 0.100$_{(11)}$    & 0.491$_{(14)}$ & 0.412$_{(14)}$ & 0.751$_{(14)}$ &     12.3     \\
NBeats      & 0.832$_{(21)}$   & 0.235$_{(9)}$ & 0.370$_{(14)}$      & 0.420$_{(19)}$   & 0.081$_{(3)}$    & 0.521$_{(17)}$ & 0.409$_{(13)}$ & 0.741$_{(12)}$ &    13.5      \\ \midrule
iTransformer & 0.645$_{(9)}$   & 0.217$_{(4)}$ & 0.258$_{(5)}$      & 0.340$_{(7)}$   & 0.085$_{(6)}$    & 0.431$_{(5)}$ & 0.380$_{(7)}$ & 0.710$_{(6)}$ &    6.1      \\
PatchTST    & 0.710$_{(12)}$   & 0.293$_{(11)}$ & 0.348$_{(12)}$      & 0.555$_{(22)}$   & 0.147$_{(15)}$    & 0.489$_{(13)}$ & 0.392$_{(11)}$ & 0.720$_{(8)}$ &    13      \\
FedFormer   & 0.744$_{(13)}$   & 0.317$_{(14)}$ & 0.341$_{(11)}$      & 0.347$_{(9)}$   & 0.233$_{(23)}$    & 0.484$_{(11)}$ & 0.413$_{(15)}$ & 0.762$_{(18)}$ &     14.3     \\
Autoformer  & 0.751$_{(14)}$   & 0.321$_{(15)}$ & 0.313$_{(8)}$      & 0.354$_{(11)}$   & 0.167$_{(16)}$    & 0.484$_{(11)}$ & 0.496$_{(20)}$ & 0.756$_{(15)}$ &    13.8      \\
Pyraformer  & 0.781$_{(19)}$   & 0.371$_{(21)}$ & 0.379$_{(16)}$      & 0.385$_{(17)}$   & 0.112$_{(13)}$    & 0.493$_{(15)}$ & 0.435$_{(16)}$ & 0.735$_{(11)}$ &    16      \\
Informer    & 0.757$_{(16)}$   & 0.336$_{(18)}$ & 0.383$_{(17)}$      & 0.364$_{(12)}$   & 0.192$_{(20)}$    & 0.605$_{(22)}$ & 0.542$_{(21)}$ & 0.772$_{(19)}$ &    18.1      \\
Transformer & 0.765$_{(17)}$   & 0.321$_{(15)}$ & 0.405$_{(19)}$      & 0.370$_{(13)}$   & 0.178$_{(19)}$    & 0.567$_{(20)}$ & 0.592$_{(22)}$ & 0.785$_{(22)}$ &   20       \\ \midrule
SCINet      & {\bf 0.601}$_{(1)}$ & {\bf0.193}$_{(1)}$ & 0.280$_{(6)}$  & 0.344$_{(8)}$   & 0.137$_{(17)}$    & 0.463$_{(9)}$ & 0.389$_{(9)}$ & 0.732$_{(9)}$ &    7.5      \\
NLinear     & 0.636$_{(6)}$   & 0.223$_{(7)}$ & {\bf 0.239}$_{(1)}$      & 0.328$_{(4)}$   & 0.091$_{(9)}$    & \uline{0.418}$_{(2)}$ & 0.375$_{(5)}$ & 0.706$_{(5)}$ &    4.9      \\
DLinear     & 0.640$_{(7)}$   & 0.497$_{(24)}$ & 0.336$_{(10)}$      & 0.444$_{(20)}$   & 0.102$_{(12)}$    & 0.442$_{(8)}$ & 0.378$_{(6)}$ & 0.686$_{(3)}$ &     8.8     \\
LSTMa       & 0.974$_{(24)}$   & 0.305$_{(13)}$ & 0.444$_{(19)}$      & 0.501$_{(21)}$   & 0.534$_{(24)}$    & 0.782$_{(24)}$ & 0.699$_{(24)}$ & 0.897$_{(24)}$ &     21.6     \\ \bottomrule
\end{tabular}}\vspace{-3mm}
\end{center}
\end{table*}
\vspace{-1mm}
\section{Experiments}

\vspace{-1mm}
{\subsection{Baselines and Experimental Settings}}
We assess the effectiveness of our model on eight datasets for MTS forecasting, namely ETTh1, ETTm1, NorPool, Caiso, Wind, Weather, Electricity, and Exchange-rate \citep{shen2024multiresolution}. Data preprocessing follows the approach outlined in previous work \citep{shen2024multiresolution}. We employ Mean Absolute Error (MAE) and Mean Squared Error (MSE)\citep{shen2020timeseries}, to measure the performance of MTS forecasting models.
We conduct time series prediction experiments by comparing 23 recent strong prediction models. 
The summary details are provided in Appendix \ref{A}.

\subsection{Main Results}
Table \ref{tab:forecasting_result} showcases the comprehensive prediction performance, with the best results highlighted in boldface and the second-best results underlined. Evaluation results indicate that our proposed method outperforms other state-of-the-art approaches in most settings. Overall, its average ranking is better than all other baselines (which include the most recent diffusion models). We report the results and the standard deviation of our model performance under five runs with different random seeds in Table \ref{tab:forecasting_result}, which exhibits that the performance of Bim-Diff is stable.

\subsection{Ablation Study}

\begin{table*}[t]
\caption{Ablation study on three datasets and the best results are highlighted in bold.}
	\label{tab:ablation_module}
	\begin{center}
		\small
        \setlength{\tabcolsep}{5.mm}
         \resizebox{1\hsize}{!}{
\begin{tabular}{ccccccccccc}
\toprule
Dataset                                                                                                           &                          & \multicolumn{3}{c}{ETTh1}                        & \multicolumn{3}{c}{Wind}          & \multicolumn{3}{c}{Weather}                      \\ \midrule
\multicolumn{1}{c|}{predict\_length}                                                                              & \multicolumn{1}{c|}{}    & 48             & 168            & 672            & 48        & 168       & 672       & 48             & 168            & 672            \\ \midrule
\multicolumn{1}{c|}{\multirow{2}{*}{ours}}                                                                        & \multicolumn{1}{c|}{MSE} & \textbf{0.339} & \textbf{0.408} & \textbf{0.456} & \textbf{0.600} & \textbf{0.893} & \textbf{1.269} & \textbf{0.120} & \textbf{0.185} & \textbf{0.305} \\
\multicolumn{1}{c|}{}                                                                                             & \multicolumn{1}{c|}{MAE} & \textbf{0.372} & \textbf{0.414} & \textbf{0.461} & \textbf{0.490} & \textbf{0.659} & \textbf{0.819} & \textbf{0.151} & \textbf{0.224} & \textbf{0.320} \\ \midrule
\multicolumn{1}{c|}{\multirow{2}{*}{w/o semantic}}                                                               & \multicolumn{1}{c|}{MSE} & 0.341          & 0.413          & 0.465          &     0.609      &     0.904      &      1.281     & 0.128          & 0.195          & 0.310
\\
\multicolumn{1}{c|}{}                                                                                             & \multicolumn{1}{c|}{MAE} & 0.370          & 0.422          & 0.472          &    0.503       &     0.664      &    0.835       & 0.159          & 0.235          & 0.329          \\ \midrule
\multicolumn{1}{c|}{\multirow{2}{*}{w/o episodic}}                                                                & \multicolumn{1}{c|}{MSE} & 0.342         & 0.414          & 0.460          &    0.610       &    0.899       &    1.275       & 0.125          & 0.187          & 0.306          \\
\multicolumn{1}{c|}{}                                                                                             & \multicolumn{1}{c|}{MAE} & 0.371          & 0.423          & 0.465          &    0.504       &     0.659     &     0.830      & 0.150          & 0.227          & 0.325          \\ \midrule
\multicolumn{1}{c|}{\multirow{2}{*}{w/o both}}                                                                    & \multicolumn{1}{c|}{MSE} & 0.352          & 0.430          & 0.477          &    0.627       &   0.930        &     1.298      & 0.136          & 0.206          & 0.322          \\
\multicolumn{1}{c|}{}                                                                                             & \multicolumn{1}{c|}{MAE} & 0.386          & 0.437          & 0.481          &   0.525       &     0.679      &     0.853      & 0.170          & 0.249          & 0.346          \\ \midrule
\multicolumn{1}{c|}{\multirow{2}{*}{w/o shared memory}}                                                           & \multicolumn{1}{c|}{MSE} & 0.348         & 0.420          & 0.470          &   0.605        &   0.905        &    1.285       & 0.121          & 0.188         & 0.310          \\
\multicolumn{1}{c|}{}                                                                                             & \multicolumn{1}{c|}{MAE} & 0.379          & 0.427         & 0.476          &     0.495      &    0.662       &    0.844      & 0.160          & 0.235          & 0.331          \\ \bottomrule
\end{tabular}}\vspace{-3mm}
\end{center}
\end{table*}

{\bf{Effect of brain-inspired memory:}} In our model, we have two key memories: semantic memory and episodic memory. We conducted an ablation study on the ETTh1, Wind, and Weather datasets, as presented in Table \ref{tab:ablation_module}. Here, "w/o" indicates the absence of a particular module, and "w/o both" refers to the conditional diffusion model without any memory modules, serving as the baseline. In the table, "ours" denotes the BimPGM without any ablations. We analyze the results shown in Table \ref{tab:ablation_module}.

1) Comparing row 4 (baseline) with rows 2 and 3, it's evident that both the semantic memory and the episodic memory contribute significantly to our model's performance. This demonstrates that both modules effectively utilize temporal patterns recurring across different channels, thereby enhancing the representation capacity of the conditional diffusion model.

2) Comparing the results on the different datasets, it can be seen that the semantic memory exhibits greater universality than the episodic memory, and this alignment with our design intention is expected. Specifically, semantic memory is designed to summarize general temporal patterns across all channels. In contrast, episodic memory focuses on remembering the special patterns across different channels that are challenging to predict. 

3) Comparing row 1 (Bim-Diff) with rows 2 and 3, it's evident that the combined memory improves model performance. This reaffirms that the two memories can indeed provide effective recurrent temporal patterns from distinct perspectives: general temporal patterns and special temporal patterns.

{\bf{w/o shared memory update:}} In this experiment, we design a brain-inspired memory for each channel independently instead of using a shared memory across different channels. This approach deviates from the shared memory update strategy proposed in this paper. The primary aim of this experiment is to assess the shared memory's capability to capture recurrent temporal patterns across different channels. This study is labeled as "w/o shared-memory." It's worth noting that comparing the results in rows 1 and 5 in Table \ref{tab:ablation_module} validates that our global shared memory module effectively captures the potential recurrent characteristics of temporal patterns across different channels.

\subsection{Analysis of Recurrent Patterns and Channel Correlations}

\begin{figure*}
\includegraphics[width = 14cm]{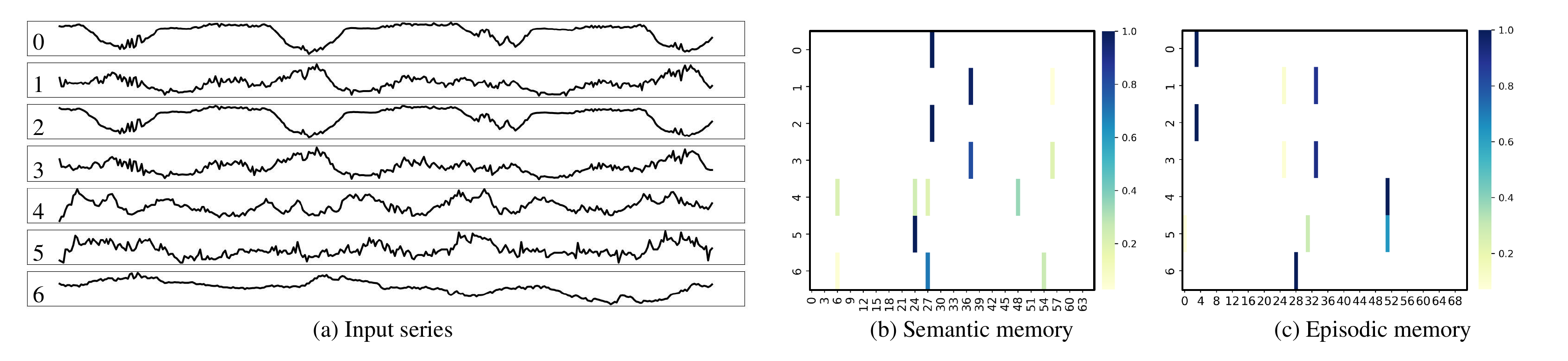}
\caption{ {The visualization of input series, similar score matrix of semantic memory and episodic memory. The input series contains 7 channels. The size of the episodic memory is set to 70, and the size of the semantic memory is set to 64.}}\label{fig:visualization}
\end{figure*}

To further validate the efficacy of our Brain-inspired Memory in capturing recurrent temporal patterns across diverse channels, we conducted experiments visualizing memory attention scores using the ETTm1 dataset, as depicted in Fig \ref{fig:visualization}. The results in the figure demonstrate that distinct channels recall relevant memories with notable similarity. This phenomenon demonstrates that our memory module can capture temporal patterns effectively, and these patterns recur among different channels. Furthermore, we suppose that the attention score of the brain-inspired memory provides an interpretable and quantifiable method for evaluating correlations between channels. For instance, we can analyze channel correlations, such as those between channel 0 and channel 2, by examining the attention score distribution across different memories. Based on the attention score, it is reasonable for the neural network to infer that channel 0 exhibits strong correlations with channel 2.

\subsection{Efficiency Analysis}

Fig~\ref{fig:c1} shows the inference time of different kinds of diffusion models. As can be seen, the inference of Bim-Diff is more efficient than mr-diff and Time-diff. The inference efficiency of Bim-Diff over existing diffusion models is due to: (i) We replace the U-Net structure in the diffusion model with a simple MLP structure. (ii) According to TimeGrad \citep{rasul2021autoregressive},  ten forward process steps are enough for diffusion to forecast. Thus, with the help of the acceleration technique DDIM \citep{song2020denoising}, which further reduces the denoising steps to just one for our method. Consequently, our model can be trained and tested on a single Nvidia 3090 GPU effectively.

\subsection{Effect of Hyper-parameters}

We evaluate the effect of two hyper-parameters:  block number of semantic memory and block number of episodic memory on the ETTh1 dataset.

\begin{figure} [t!]\vspace{-3mm}
	\centering
	\subfigure[\label{fig:a1}]{
		\includegraphics[scale=0.22]{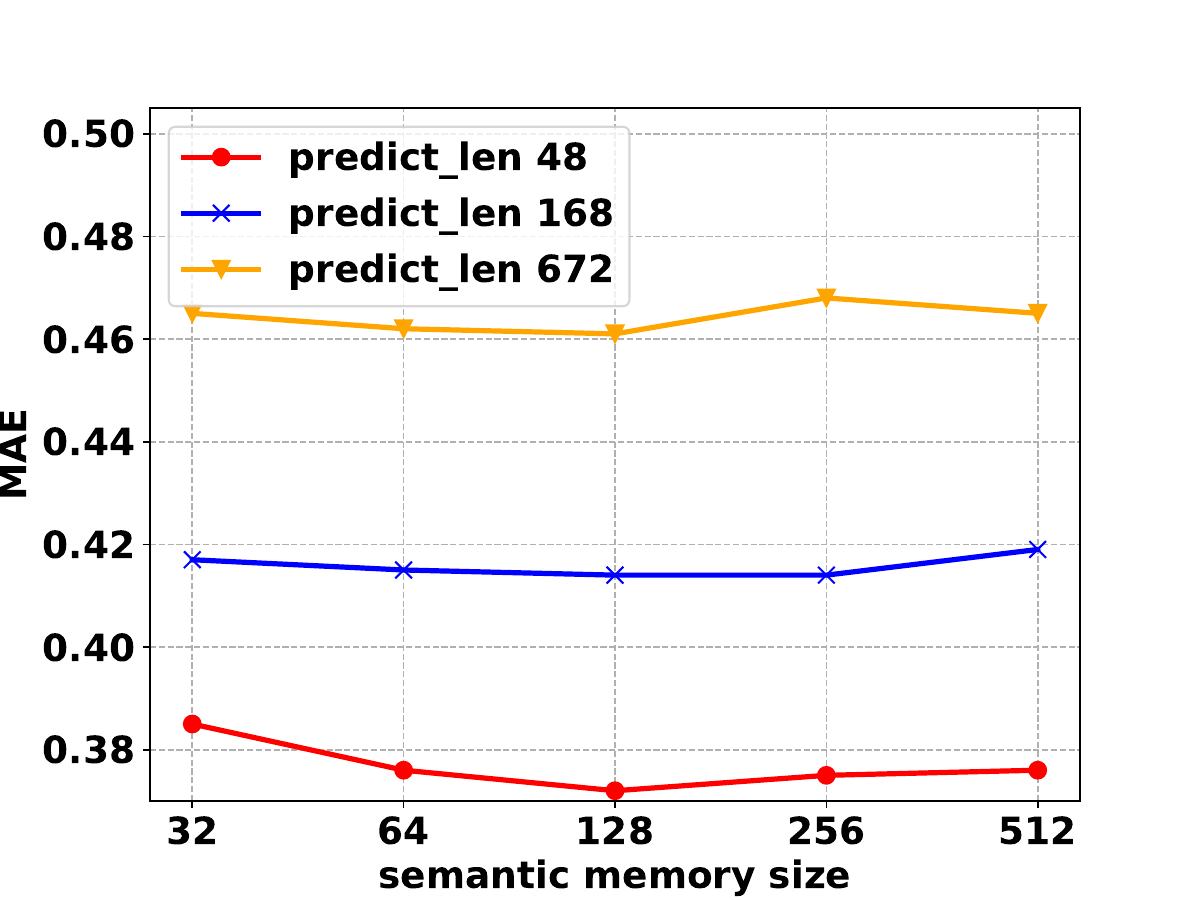}}
	\subfigure[\label{fig:b1}]{
		\includegraphics[scale=0.22]{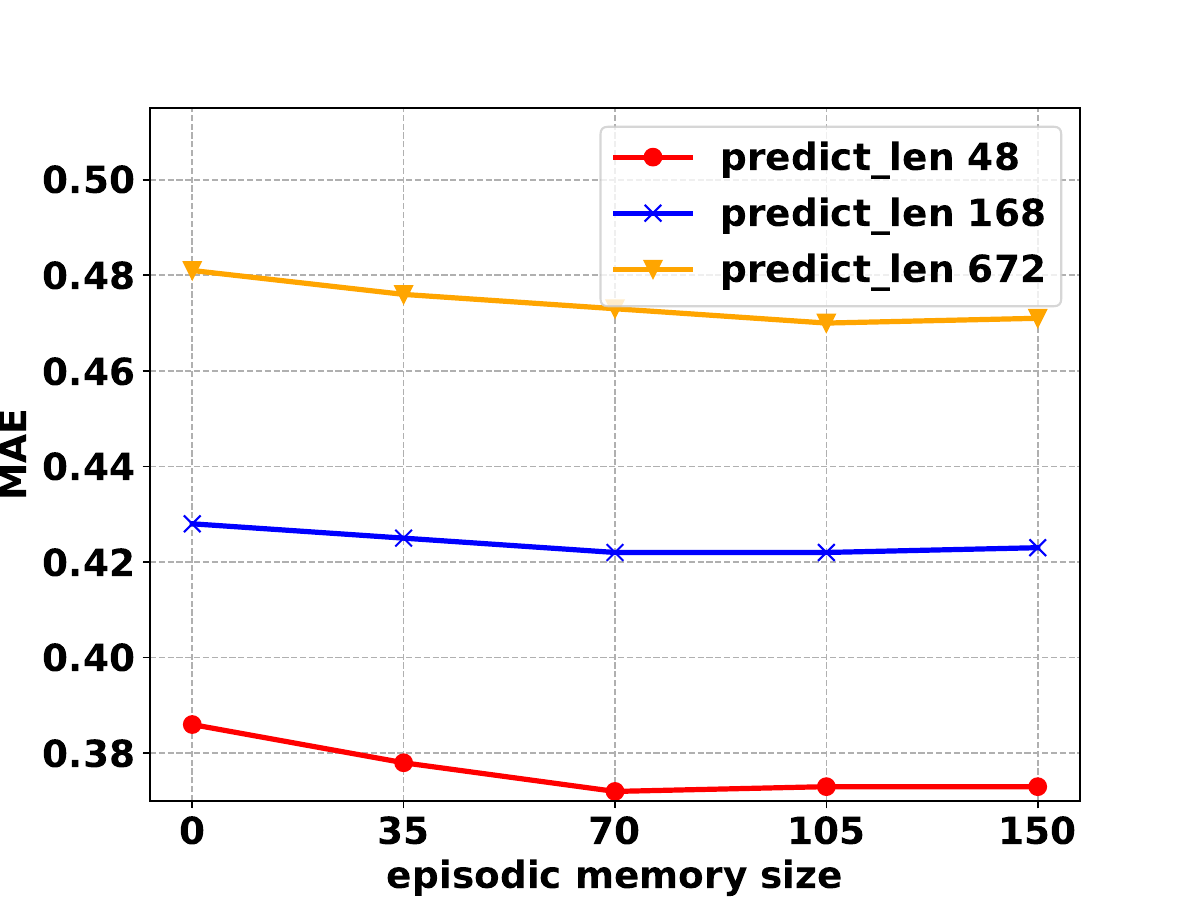}}
  	\subfigure[\label{fig:c1}]{
		\includegraphics[scale=0.22]{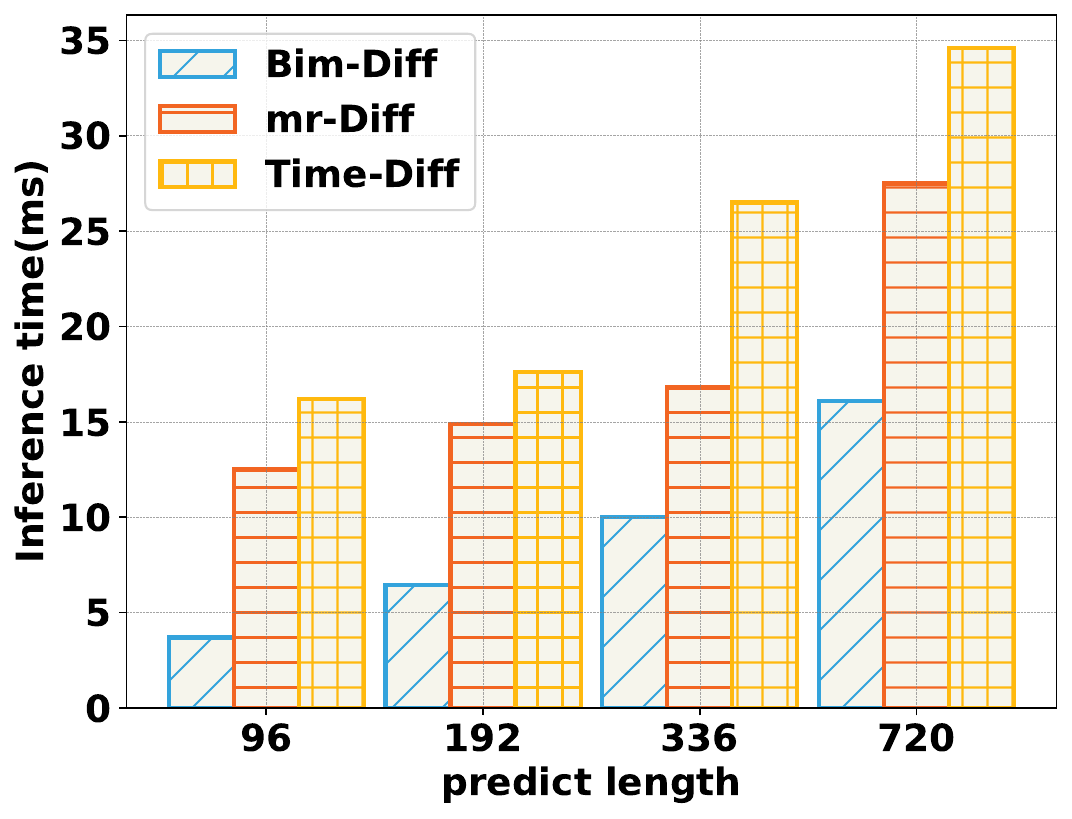}} 
	\caption{(a)-(b): The effect of the size of semantic (a) and episodic (b) memory on ETTh1 dataset. (c): Inference time (in ms) of various time series diffusion models with different prediction horizons (H) on the ETTh1(other method results come from \citep{shen2024multiresolution}).}\vspace{-3mm}
	\label{paremeter_vis} 
\end{figure}

{\bf{Effect of memory size on semantic memory:}} In Fig.\ref{fig:a1}, we set the memory size $N_1$ from 32 to 512 and evaluate MSE with different prediction lengths on ETTh1 dataset. To emphasize the impact of the memory size of the semantic memory, we conducted experiments without the episodic memory module. It becomes apparent that the semantic memory size $N_1$ follows a non-linear relationship with performance. An excessively large memory size can be detrimental because the semantic memory must learn to summarize patterns independently. In such cases, it might struggle to effectively capture useful patterns. Conversely, when the memory size is too small, it may not have the capacity to capture the essential patterns efficiently. Finding an optimal memory size is essential for striking the right balance between semantic retention and model efficiency.

{\bf{Effect of memory size on episodic memory:}} In Fig.\ref{fig:b1}, the memory size $N_2$ is set from 0 to 140 and evaluates MSE with different predicted lengths on ETTh1 dataset. In order to highlight the effect of the memory size of the episodic memory, we remove the semantic memory. The results in Fig.\ref{fig:b1} demonstrate that the performance tends to improve as the memory size increases. However, there is a point at which the performance gains start to diminish. This phenomenon underscores the effectiveness of the episodic memory we designed. This module stores special patterns for each channel, and when the memory size is sufficiently large, it can record all useful temporal patterns.
\vspace{-2mm}
\section{Conclusions}
\vspace{-1mm}
In this paper, we present a novel approach named Bim-Diff designed to capture general and special temporal patterns across different channels in multivariate time series forecasting tasks. Bim-Diff consists of two fundamental components: semantic memory and episodic memory, each capturing recurrent temporal patterns from different perspectives. Bim-Diff demonstrates excellence in extracting robust and expressive representations from MTS, leading to superior performance compared to other models in MTS forecasting tasks. Additionally, due to our effective simplification, the diffusion model for MTS forecasting can be trained efficiently. Empirical results from various MTS forecasting experiments provide compelling evidence of the effectiveness of our proposed model.


\normalem
\bibliography{neurips2024}
\bibliographystyle{unsrtnat}

\newpage
\appendix


\section{Details of Experiments}
\label{A}
\subsection{Datasets statistics}

The details of the datasets: (1) ETT dataset includes the time series of oil de-stationary factors and power load collected by electricity transformers from July 2016 to July 2018.
ETTm1 is recorded every 15 minutes, and ETTh1 is recorded every 60 minutes. 
(2) The exchange dataset includes the panel data of daily exchange rates from 8 countries from 1990 to 2016.
(3) The ILI dataset collects the ratio of influenza-like illness patients versus the total patients in one week,
which is recorded weekly by the Centers for Disease Control and Prevention of the United States from
2002 and 2021.
(4) Weather dataset includes meteorological time series with 21 weather indicators collected from the Weather Station of the Max Planck Biogeochemistry Institute in 2020 every 10 minutes.
(5) The Electricity dataset includes the electricity consumption of 321 customers recorded hourly from 2012 to 2014.
(6) NorPool dataset includes eight years of hourly energy production volume series in multiple European countries; 
(7) The Caiso dataset contains eight years of hourly actual electricity load series in different zones of California;
(8) The Wind dataset contains wind power records from 2020-2021 at 15-minute intervals
\begin{table}[h]
\begin{center}
\caption{Summary of statistics of datasets}
\label{tab:dataset statistics}
\begin{tabular}{ccccccc}
\hline
Datasets & Samples & channel number & Sample Rate  & steps(H)\\ \toprule
ETTh1  & 17420   & 7   & 60 min &168 (1 week) \\ \midrule
ETTm1  & 69680   & 7   & 15 min &192  (2 days)\\ \midrule
Exchange  & 7588   & 8   & 1 day &14 (two weeks) \\ \midrule
Wind    & 48673   & 7   & 15 min & 192(2 days)\\ \midrule
Norpool  & 70128   & 18   & 60 min  &720 (1 month) \\ \midrule
Caiso  & 74472   & 10   & 60 min &720 (1 month)\\ \midrule
Weather  & 52695   & 21   & 10 min  & 672 (1 week)\\ \midrule
Electricity  & 26304   & 321   & 60 min & 168 (1 week)\\ \bottomrule
\end{tabular}
\end{center}
\end{table}

\subsection{Evaluation metrics}

We use two evaluation metrics which is usually used in time series forecasting tasks to measure the performance of predictive models. Let $\Xv_{:,i} \in \mathbb{R}^{N \times 1}$ be the ground truth data of all channels at time step i, $\Xv_{:,i}^{\prime} \in \mathbb{R}^{N \times 1}$ be the predicted values, and $\Omegav$ be indices of observed samples. The metrics are defined as follows.

Mean Absolute Error (MAE)
\begin{equation}
M A E=\frac{1}{|\Omega|} \sum_{i \in \Omega}\left|\boldsymbol{X}_{:, i}-\boldsymbol{X}_{:, i}^{\prime}\right|
\end{equation}

Mean Square Error (MSE)
\begin{equation}
M S E=\frac{1}{|\Omega|} \sum_{i \in \Omega}\left|{\boldsymbol{X}_{:, i}-\boldsymbol{X}_{:, i}^{\prime}}\right|^2
\end{equation}
\subsection{Baseline methods}
The details of the baselines are as follows: The input lengths for all baseline experiments are chosen from $\{96, 336, 720, 960\}$, and the reported results are based on selecting the best-performing outcome within this range.

{\bf{iTransformer}}: iTransformer is reproduced using the original paper’s configuration in the official code or quoted from their original paper. 

Other methods' results are cited from \citep{shen2024multiresolution}.

\subsection{Implementation details}
{\bf{Our model}}: All the experiments are implemented with PyTorch and conducted
on a single NVIDIA $3090$ 24GB GPU. Each model is trained by an ADAM optimizer using MSE loss. The input length for the Bim-Diff is chosen from $(96, 336, 720, 960)$. For the illness dataset, the input length is set to 14. The splitting ratio is set to 7:1:2 for train, val, and test set on the illness, weather, exchange, and Electricity dataset, which is the same as previous work. The splitting ratio is set at 3:1:2 for train, val, and test set on ETTh1 and ETTm1 datasets, which is the same as previous work.

\section{Additional Model Analysis}
\subsection{Update strategy analysis}
{\subsubsection{Episodic memory}}
\label{B.1.1}

\begin{figure}[h]
\includegraphics[width = 14cm]{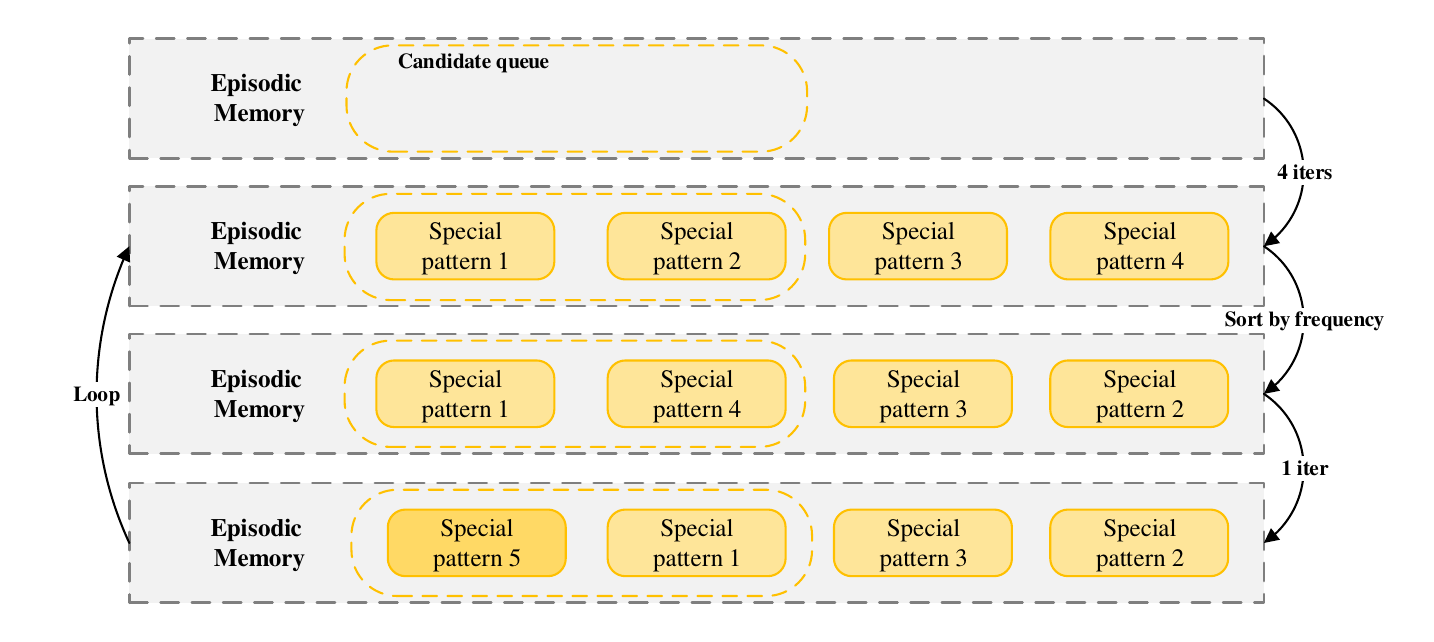}
\caption{ {The details of episodic memory update strategy.}}\label{fig:episodic}
\end{figure}

In our approach, we make an initial assumption that a new special pattern should be incorporated into the memory after each iteration. At the outset, the episodic memory is empty. As the memory fills up, we implement a sorting mechanism based on access frequency. Notably, only the last pattern in the candidate queue participates in this sorting process, as illustrated in Figure \ref{fig:episodic}. When a new special pattern arrives, the last element in the candidate queue is popped out. Subsequently, the new pattern pops in the candidate queue. This strategy effectively addresses the memory solidification issue that we previously discussed.
{\subsubsection{Semantic memory}}
\label{B.1.2}
\begin{figure}[h]
\includegraphics[width = 14cm]{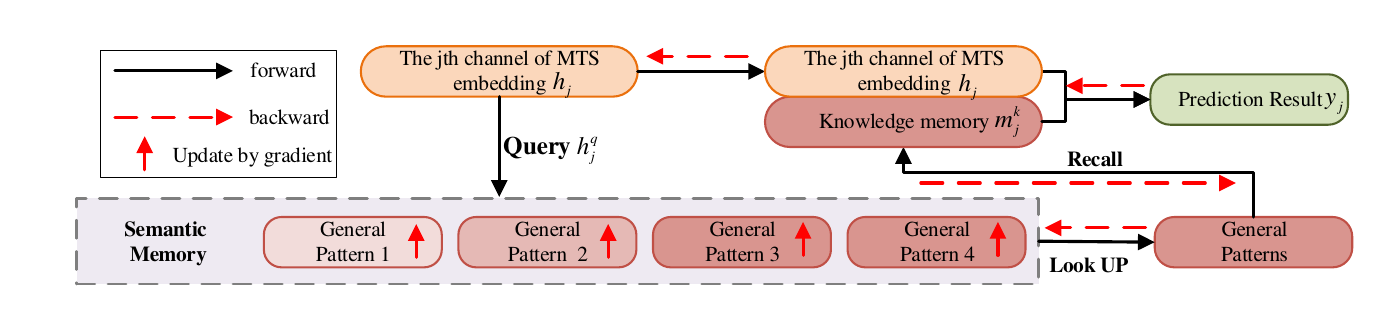}
\caption{ {The details of semantic memory update strategy.}}\label{fig:knowledge}
\end{figure}

The specific update process of the semantic memory is elaborated in Figure \ref{fig:knowledge}. Since each general pattern contributes to predicting the $j$th channel of the MTS $\yv_j$, every pattern undergoes updates during the backward process and accumulates information from the $j$th channel of the MTS. 
Because all general patterns are shared when utilized to forecast the outcomes of each channel, each pattern aggregates information from every channel, enabling it to capture general patterns across different channels effectively.

\subsection{Complexity analysis}
We have added computational complexity of the proposed approach compared to existing methods as shown below: The original Transformer has $O(L^2)$ complexity on both time and space, where L is the number of input lengths. iTransformer uses an inverted Transformer and achieves complexity $O(D^2)$ on both time and space, where D is the number of channels. Dlinear is a linear-based model, so the complexity of both time and space is $O(L)$. Because we replace the U-Net in the traditional diffusion model with an MLP, the complexity of our model is $O(DM)$ mainly caused by the calculation of attention score, which is (D is the number of channels, M is the memory size). In most cases, D, M is much smaller than L, so the complexity of our model is smaller than the transformer-based model.

\end{document}